\documentclass[10pt,twocolumn,letterpaper]{article}

\usepackage{wacv}              

\usepackage{graphicx}
\usepackage{amsmath}
\usepackage{amssymb}
\usepackage{booktabs}
\usepackage{multirow}
\usepackage{enumitem}
\usepackage{siunitx}
\usepackage{soul}
\usepackage[accsupp]{axessibility}  

\usepackage[pagebackref,breaklinks,colorlinks]{
hyperref}

\usepackage[capitalize]{cleveref}
\crefname{section}{Sec.}{Secs.}
\Crefname{section}{Section}{Sections}
\Crefname{table}{Table}{Tables}
\crefname{table}{Tab.}{Tabs.}

\def\wacvPaperID{1218} 
\def\confName{WACV}
\def\confYear{2026}

\begin{document}

\title{Diffusion-Based Authentication of Copy Detection Patterns: A Multimodal Framework with Printer Signature Conditioning} 

\author{
  Bolutife Atoki \hspace{2em} 
  Iuliia Tkachenko \hspace{2em}
  Bertrand Kerautret \hspace{2em}
  Carlos Crispim-Junior\\
Université Lumière Lyon 2, CNRS, INSA Lyon, Universite Claude Bernard Lyon 1, LIRIS\\
UMR5205, 69007 Lyon, France\\
{\tt\small \{bolutife.atoki, iuliia.tkachenko, bertrand.kerautret, 	carlos.crispim-junior\}@liris.cnrs.fr}
}
\maketitle

\begin{abstract}\footnote{This paper has been accepted at WACV 2026.}
Counterfeiting affects diverse industries, including pharmaceuticals, electronics, and food, posing serious health and economic risks. Printable unclonable codes, such as Copy Detection Patterns (CDPs), are widely used as an anti-counterfeiting measure and are applied to products and packaging. However, the increasing availability of high-resolution printing and scanning devices, along with advances in generative deep learning, undermines traditional authentication systems, which often fail to distinguish high-quality counterfeits from genuine prints. In this work, we propose a diffusion-based authentication framework that jointly leverages the original binary template, the printed CDP, and a representation of printer identity that captures relevant semantic information. Formulating authentication as multi-class printer classification over printer signatures lets our model capture fine-grained, device-specific features via spatial and textual conditioning. We extend ControlNet by repurposing the denoising process for class-conditioned noise prediction, enabling effective printer classification. 
On the Indigo \(1\times1\) Base dataset, our method outperforms traditional similarity metrics and prior deep learning approaches. Results show the framework generalizes to  counterfeit types unseen during training.
\end{abstract}

\section{Introduction}
\label{sec:intro}
Counterfeiting impacts industries from pharmaceuticals to food, posing serious health, safety, and economic risks. The widespread availability of high-quality printing and scanning devices, coupled with advances in generative deep learning methods, has driven the increase in counterfeit products. To address this challenge, some anti-counterfeiting solutions employ printable unclonable codes applied to products or packaging. A simple and cost-effective form of these codes is the Copy Detection Pattern (CDP), a maximum-entropy image that is sensitive to copying and undergoes information loss when reprinted\cite{2004SPIE.5310..176P}.

\begin{figure}[t]
    \centering
    \includegraphics[width=1\linewidth]{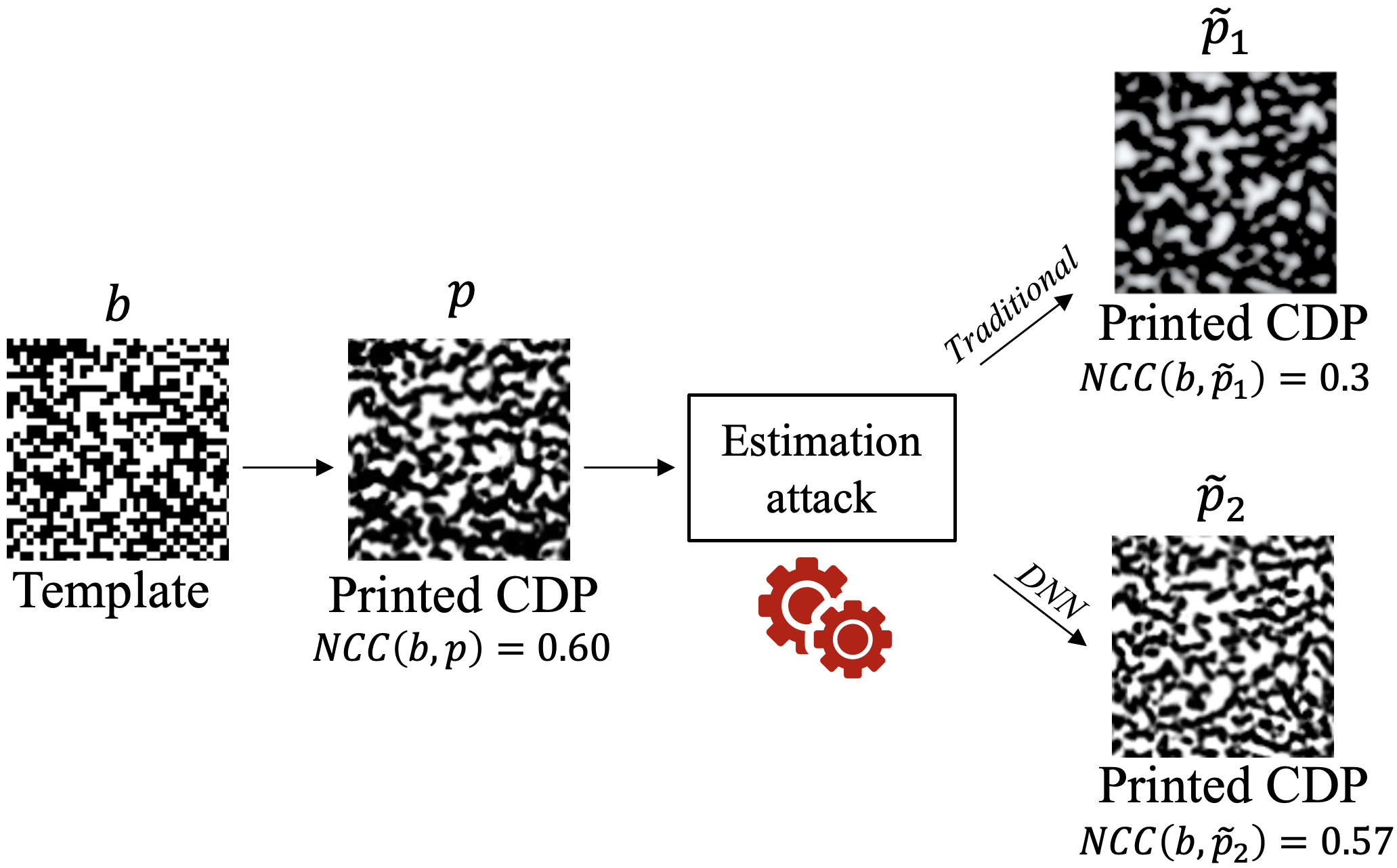}
    \caption{Examples of a binary template, authentic printed CDP, and counterfeit printed CDPs from traditional image processing and deep learning methods. Normalised cross-correlation reliably detects traditional counterfeits but is unreliable against deep learning-based ones.}
    \label{fig:illustration}
\end{figure}

However, advances in generative methods have introduced new challenges. Modern computer vision approaches\cite{DBLP:conf/nips/GoodfellowPMXWOCB14, DBLP:conf/miccai/RonnebergerFB15} enable the estimation of binary approximations of CDPs, facilitating near-perfect replicas with little information loss, and undermine the reliability of traditional authentication methods\cite{ DBLP:conf/wifs/YadavTTF19, Chaban2021MachineLA} as illustrated in \Cref{fig:illustration}. In response, recent approaches have examined authentication using the binary template alone, printed CDP alone, or a combination of both.

Independent studies\cite{4806207, 8726322, 5537996, 4ce75c4277c74bb2939985b344fb7735} show printers leave unique, repeatable signatures on printed outputs due to subtle hardware and mechanical variations. These printer-specific signatures capture the identity of the printer, enabling discrimination even between different units of the same model. Unlike prior work using printer information only for grouping or evaluation, we treat the printer as an identity-bearing entity central to authentication. We represent each printer's identity using a human-readable textual description, enabling the model to learn associations between signature patterns and printer classes, thereby supporting broader generalisation.

We introduce a unified authentication framework that integrates the binary template, printed CDP, and printer identity in a single decision-making process. We formulate authentication as a multi-class classification task, where the goal is to identify the source printer. Our framework operates in the latent space using a diffusion-based architecture, which enables the model to capture low-dimensional representations that retain fine-grained, printer-specific details. Compared to prior generative approaches, Generative Adversarial Networks (GANs) \cite{DBLP:conf/nips/GoodfellowPMXWOCB14} produce sharp images but often suffer from mode collapse and reduced diversity \cite{DBLP:conf/icml/Arora0LMZ17}, while Variational Autoencoders (VAEs) \cite{DBLP:journals/corr/KingmaW13} cover more modes but typically yield blurrier results\cite{DBLP:conf/iclr/BredellFCEK23}; diffusion models achieve both high fidelity and broad distribution coverage \cite{10.5555/3540261.3540933}. A textual description of the printer and a spatial conditioning mechanism guide the model’s attention to regions likely to contain these signature patterns. To perform classification, we introduce a novel classification block within a semantically and spatially conditioned setting, allowing the model to extract the printer’s signature from both the binary template and the printed CDP. Authentication is achieved by comparing the predicted printer identity with the expected one associated with the authentic CDP. Discrepancies between the two indicate potential counterfeiting.

In summary, our contributions are four-fold. First, we incorporate printer-specific signatures into the CDP authentication process, treating the printer as an identity-bearing entity rather than auxiliary metadata. Second, we propose the first unified authentication framework that jointly considers the binary template, printed CDP, and printer identity within a single pipeline. Third, we enable cross-printer generalisation by formulating authentication as a multi-class classification task over printer identities. Finally, we extend the image-conditioned diffusion model ControlNet\cite{DBLP:conf/iccv/ZhangRA23} beyond its generative purpose by modifying its architecture to support classification. Specifically, we introduce a classification module that leverages the model’s learned denoising capabilities to perform class discrimination. Our implementation and models are available at our public repository \footnote{Repository URL is available at \url{https://gitlab.liris.cnrs.fr/anr-trustit/diffusion_based_cdp_authentication.git}}.

The remainder of this paper is as follows. In \Cref{sec:lit-review}, we review existing work on CDP authentication. \Cref{sec:methodology} describes the proposed methodology and authentication strategy. In \Cref{sec:experiments}, we perform experiments and the evaluation of our method. \Cref{sec:conclusion} concludes the paper with a summary of our findings and perspectives for future research.

\section{Previous Work}
\label{sec:lit-review}
In this section, we review existing methods for authenticating copy detection patterns (CDPs). We begin by briefly discussing common counterfeiting techniques used to attack authentication systems, as understanding these attacks is essential for evaluating the effectiveness of various authentication approaches. Then, we broadly categorize existing authentication methods into three main approaches based on the type of data they utilise: (i) template-based methods, (ii) printer type-based methods, and (iii) printed CDP-based authentication methods.

Counterfeiters estimate binary templates using techniques from thresholding (e.g., Otsu) to deep learning. This process is illustrated in \Cref{fig:illustration}, which shows the progression from a genuine binary template to an authentic printed CDP, followed by template estimation attacks and their printing as counterfeits. Models like Bottleneck DNN, autoencoders, SRGAN, and U-Net produce high-resolution approximations closely mimicking authentic templates \cite{8682967, 10.1145/3335203.3335718, DBLP:conf/wifs/YadavTTF19, Chaban2021MachineLA}. Given the variety and sophistication of these counterfeiting techniques, authentication methods have been developed to detect and prevent forgery. These methods can be broadly classified based on the type of data they utilise, as detailed below.

\inlineheading{Template-based CDP authentication} The simplest form of template-based authentication compares a test printed CDP (probe) to its corresponding binary template using similarity metrics such as Normalised Cross-Correlation (NCC) or Structural Similarity Index (SSIM). A similarity score is computed and evaluated against a threshold derived from authentic pairs. These methods once sufficed but now fail against high-quality template estimates \cite{DBLP:conf/wifs/BelousovPCTTHV22, Chaban2021MachineLA, DBLP:conf/wifs/YadavTTF19}.

To improve reliability, more advanced approaches have been proposed. The authors in\cite{DBLP:conf/wifs/TuttTCPBHV22} introduced a method that models the printing-and-digitisation (P\&D) process as a transformation of local \(3\times3\) binary neighbourhoods. For each printer, a codebook is constructed, mapping the probability of bit-flip for every one of the 512 possible neighbourhood configurations. At authentication, the flipped bits in the binarised probe are compared to the expected flipping probabilities, and authentication is performed via thresholding. Later, Tutt \etal\cite{DBLP:journals/tifs/TuttTCPBHV24} extended this idea by using a two-class SVM trained on the bit-flip probabilities and the corresponding centre pixel values for each neighbourhood.

Though these methods surpass basic similarity metrics, they only indirectly capture P\&D noise through empirical bit-flip probabilities. The noise is not explicitly modelled or integrated into the learning process, which limits adaptability. Consequently, a separate codebook is required for each printer, preventing generalisation across devices and restricting scalability in real-world deployments.

\inlineheading{Printer type-based authentication} These methods rely solely on the image of the printed CDP for authentication, based on the assumption that printed outputs exhibit characteristics unique to their printer type. One such method is proposed by Zeghidi \etal\cite{DBLP:conf/wifs/ZeghidiCT23}, who use a Siamese Neural Network\cite{DBLP:conf/cogsci/RaoWC16} to embed printed CDP pairs into a shared feature space. The network is trained to minimise the distance between embeddings from the same printer and maximize it for different printers. At test time, a probe is compared to multiple reference samples from the claimed printer, and is accepted as authentic if the majority of distances fall below a learned threshold. Because it ignores the template, this method may fail when unknown templates are printed on known printers.

\inlineheading{Printed CDP-based authentication} Authentication methods that compare the test printed CDP with its authentic counterpart have proved to yield reliable authentication performance\cite{DBLP:conf/wifs/PulferBTCTHV22}. However, directly comparing a probe to its corresponding authentic sample is costly and impractical at scale due to the need for storing or retrieving large databases of reference prints. To address this, generative methods are employed to synthesize the expected appearance of a printed CDP directly from its digital template, enabling on-demand reference generation and eliminating the need for physical sample storage. This enables the authentication process to rely on comparisons with a predicted print, thereby reducing overhead and improving scalability.

Pulfer \etal\cite{DBLP:conf/wifs/PulferBTCTHV22} employed a U-Net model to predict a print from a given binary template, leveraging its effectiveness in image-to-image tasks. The authentication decision was based on the masked pixel-wise error between the probe and the predicted print, under the assumption that an authentic printed CDP would closely match the predicted print. Chaban \etal\cite{DBLP:conf/wifs/ChabanPV24} adopted a Pix2Pix framework\cite{DBLP:conf/cvpr/IsolaZZE17}, which utilizes a conditional GAN to enhance realism and fidelity in the synthesized outputs. The adversarial training in Pix2Pix encourages the generator to produce outputs indistinguishable from real printed samples, making it more robust to subtle printing variations. The Pearson Correlation Coefficient (PCC) was used as a similarity metric, emphasizing global structural alignment between the synthesized and probe CDPs. However, while these methods can generate plausible printed CDP images for new, unseen templates, they are trained on samples from a single printer type, as they lack explicit conditioning on printer identity. Consequently, their outputs belong solely to the distribution of the printer they were trained on and do not handle samples from other printers.\\

While prior work has advanced CDP authentication by leveraging the template and the printed output, existing methods are typically trained on data from a single printer and lack mechanisms to explicitly model printer-specific variations. As a result, they are limited in their ability to perform reliably across different printer types, and often fail to account for the characteristic distortions introduced by specific printing processes. To overcome these limitations, we propose a novel authentication framework that, for the first time, jointly considers the binary template, printed CDP, and printer identity. The following section details our methodology for leveraging this combined information to extract printer-specific signatures and improve counterfeit detection.

\begin{figure*}[t]
    \centering
    \includegraphics[width=1\linewidth]{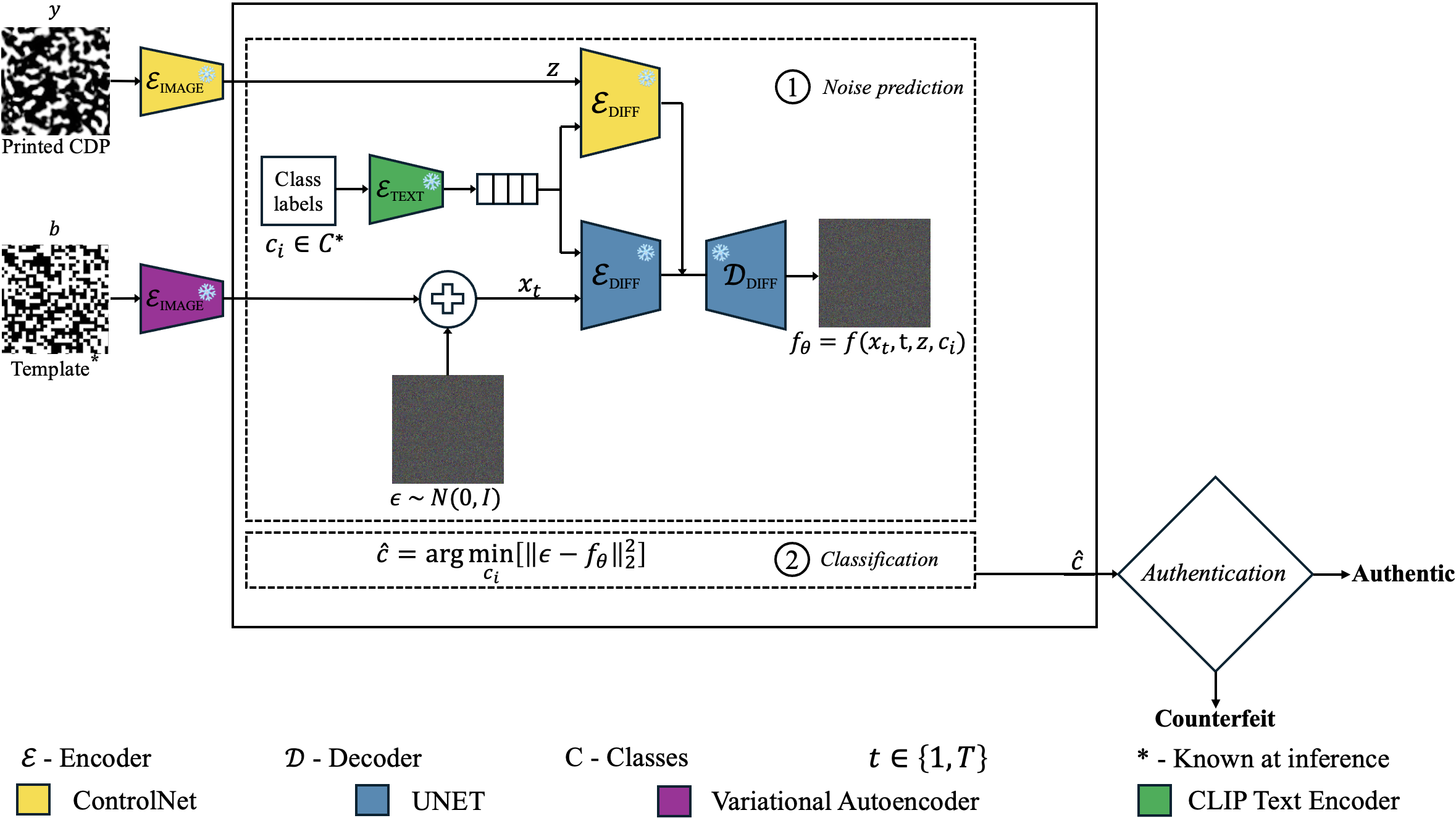}
    \caption{Authentication pipeline leveraging printer signature reconstruction. The reverse diffusion process takes a noised template as input and is conditioned on the printed CDP and printer identity text. ControlNet is modified to predict noise per class and includes a classification module selecting the class with minimal prediction error.}
    \label{fig:scheme}
\end{figure*}

\section{Methodology}
\label{sec:methodology}
Our approach to counterfeit detection is based on the printed CDP, binary template, and printer identity. The core idea is to encode these inputs into a latent representation that captures printer-specific characteristics. Authentication is then based on detecting inconsistencies between the latent signature inferred from a candidate CDP and the expected signature for the known authentic printer. To achieve this, we newly formulate authentication as a supervised multi-class classification problem. Each known authentic printer corresponds to a unique class, and various categories of counterfeit CDPs are modelled as additional classes. This formulation enables training on data from multiple printers and supports cross-printer authentication.

\subsection{Definitions}
\label{subsec:definitions}
Let the original binary template be denoted as \(b\), and the authentic CDP printed from \(b\) as \(p\). The estimated template from a printed CDP is represented as \(\tilde{b}\), while \(\tilde{p}\) denotes a counterfeit CDP printed from \(\tilde{b}\). The candidate CDP presented for authentication is denoted as 
\( y \in \{p, \tilde{p}\} \), where \( y \) may be either authentic \(p\) or counterfeit \(\tilde{p}\). 

In the diffusion framework, the template \( p \) is encoded into a latent representation \( x_0 \), which is subsequently noised to timestep \( t \), yielding \( x_t \), where \( 0 < t \leq T \). Let \( z \) denote the latent representation of the printed CDP corresponding to \( y \). Let \( \{c_i\}_{i=1}^k \) denote the set of class labels, where each \( c_i \) encodes textual information describing the printing process used to generate CDPs in class \( i \).

\subsection{Diffusion-Based Signature Classification}
\label{subsec:classification}
Our method learns printer-specific signatures by modelling the reverse diffusion process conditioned on both the printed CDP and printer identity. The strategy relies on the fact that the reverse process follows the correct denoising path only when the conditioning information aligns with the true source printer of the candidate CDP.

\paragraph{Forward diffusion:} We begin with the standard forward diffusion process, where Gaussian noise is gradually added to the template latent \( x_0 \) over \( T \) steps. Here, \( t = 0 \) signifies the original (noise-free) latent, and noise increases progressively with \( t \). 

\begin{equation}
q(x_t | x_0) = \mathcal{N}(x_t; \sqrt{\bar{\alpha}_t} x_0, (1 - \bar{\alpha}_t) \mathbf{I}),
\label{eq:forward_diffusion}
\end{equation}

where \( \bar{\alpha}_t = \prod_{s=1}^t \alpha_s \), and \( \alpha_t \in (0, 1) \) denotes the variance-preserving noise schedule controlling the amount of noise added at each timestep.

\paragraph{Reverse diffusion:} The model learns the reverse denoising process, conditioned on the latent printed CDP \( z \) and textual printer information \( c_i \), to predict the added noise \( \epsilon \). Since textual conditioning provides global semantic context associated with the printer class, it is limited in its ability to offer fine-grained spatial control. To address this, the printed CDP latent \( z \) is included as an image-based conditioning signal, guiding the model’s attention toward spatially localised features that reflect printer-specific artifacts.

\begin{equation}
f_\theta(x_t, t, z, c_i) \approx \epsilon,
\label{eq:reverse_diffusion}
\end{equation}

where \( f_\theta \) is a neural network trained to minimise the difference between predicted and actual noise.

\subsection{Framework components}
\label{subsec:framework-components}
Our framework performs signature classification by modelling the correct denoising trajectory of a noised binary template, conditioned on both printer identity and the corresponding printed CDP. Textual conditioning offers global semantics but lacks spatial specificity to capture localised printer features. To address this, recent image-conditioned diffusion models incorporate spatial alignment using cross-attention or feature injection mechanisms \cite{DBLP:conf/aaai/MouWXW0QS24, DBLP:conf/iccv/ZhangRA23}. We adopt the ControlNet architecture \cite{DBLP:conf/iccv/ZhangRA23},  which introduces spatial control in diffusion models through a parallel, trainable branch that injects conditioning (e.g., edge maps or segmentation masks) into the denoising U-Net at multiple resolution levels. In our case, we condition the model on the printed CDP, encoded into a latent representation, to guide denoising toward printer-discriminative regions.

We extend ControlNet beyond its original generative purpose by modifying its architecture to support classification. This involves two key architectural changes, highlighted in \Cref{fig:scheme}: (Block 1) the model is adapted to predict noise independently for each candidate class, retaining only the predicted noise rather than generating images, and (Block 2) a classification module is added, leveraging the model’s learned denoising capabilities to select the class with the lowest prediction error, measured as the discrepancy between predicted and actual noise. Our approach builds on the diffusion classifier paradigm introduced in \cite{DBLP:conf/iccv/LiPDBP23}, which performs zero-shot classification by identifying the class label whose conditioning minimises noise prediction error. Our model conditions the reverse diffusion process on both the text description of printer identity and the printed CDP. For a given noised latent \(x_t\), the model predicts the noise under each candidate class \( c_i \), and selects the class with the lowest error:

\begin{equation}
\hat{c} = \arg\min_{c_i} \mathbb{E}_t \left[ \| \epsilon - f_\theta(x_t, t, z, c_i) \|_2^2 \right],
\label{eq:classification}
\end{equation}

where \( \hat{c} \) is the predicted printer class, and \( \mathbb{E}_t \) denotes expectation

\Cref{fig:scheme} shows the binary template encoded and perturbed with Gaussian noise at a random timestep \( t \). The printer identity is encoded as a text embedding, and the printed CDP is encoded into an image latent. These are used to condition the model, which then predicts the noise for each candidate class. To improve robustness and reduce the impact of timestep-specific bias, this process is repeated \(N\) times with different randomly sampled timesteps. The final predicted class is selected as the one with the lowest aggregated prediction error across all sampled timesteps.

\inlineheading{Authentication strategy via classification}
Our authentication strategy compares the predicted printer class \( \hat{c} \), obtained by classifying the probe CDP using our diffusion-based model conditioned on printer identity, with the known ground-truth printer class \( c^* \), which is assumed to be available at test time. For each candidate CDP \( y \), authentication is performed by verifying whether \( \hat{c} == c^* \). A match indicates the CDP was printed by the correct authentic printer and is deemed authentic. Any mismatch implies a counterfeit, even if \( c^* \) corresponds to a different legitimate printer from the same manufacturer. This per-printer authentication provides finer granularity than manufacturer-level checks, enhancing security by preventing adversaries from evading detection by using another authorized printer. Although printers from the same manufacturer may share similar global signatures, only the authentic printer is expected to replicate the exact spatial and structural features present in the original CDP, making impersonation difficult. We assume a closed-set classification setting where the complete set of candidate printer classes, including both authentic and counterfeit printers, is known at test time. Formally, the authentication decision rule is:

\begin{equation}
\text{Auth}(y) = 
\begin{cases}
\text{Authentic}, & \text{if } \hat{c} = c^* \\
\text{Counterfeit}, & \text{otherwise}
\end{cases}
\label{eq:decision_rule}
\end{equation}

\section{Experiments}
\label{sec:experiments}

\subsection{Experimental details}
\label{subsec:experimental-details}

\inlineheading{Dataset Selection} Existing datasets for CDP authentication provide limited support for modelling printer-specific signatures. The Scantrust dataset \cite{DBLP:conf/wifs/KhermazaTP21} includes only a single printer and simulates counterfeit samples by varying binarisation methods, which does not support learning printer-dependent characteristics. The Indigo datasets offer improved support with two printers and are available in two variants: the Indigo \(1\times1\) Base dataset \cite{Chaban2021MachineLA} and the Indigo Variability dataset \cite{Chaban2022wifs}. The latter extends the Base variant by introducing four paper types to increase variability. Though limited to two printers and one scanner, these datasets are the most widely used for CDP authentication. We adopt the Indigo \(1\times1\) Base dataset to model fine-grained printer-specific signatures under controlled conditions, as it introduces hardware variability without confounding factors. It contains binary templates and corresponding printed CDPs produced by two industrial printers: HP Indigo 5500 and HP Indigo 7600.

We encode printer identity using natural language descriptions of the printing process to align with CLIP’s \cite{DBLP:conf/icml/RadfordKHRGASAM21} pre-trained text encoder. For example, an authentic print from the HP Indigo 5500 is described as: ``Data Matrix image printed with HP Indigo 5500 printer'', while a counterfeit reprinted with the 7600 is described as: ``Data Matrix image printed with HP Indigo 5500 printer, then reprinted with HP Indigo 7600 printer''. These semantically rich descriptions better reflect the natural language patterns seen during CLIP’s training, potentially improving text–image alignment compared to minimal phrases like ``HP Indigo 5500''. We reserve alternative representations of printer identity for ablation studies. Overall, this approach leverages CLIP’s language priors to leverage meaningful relationships between printer classes and to support learning of subtle, device-specific signatures through multi-printer training.

\inlineheading{Dataset reformulation for classification}
The Indigo dataset is reformulated into a six-class classification task. Two classes correspond to authentic prints from HP Indigo 5500 and HP Indigo 7600, respectively. The remaining four are counterfeit types generated by estimating templates from authentic prints and reprinting them using either of the two printers. These are labeled using the format \textit{HPXX\_YY}, where \textit{XX} refers to the original printer and \textit{YY} to the reprinting device. Each class contains $720$ samples, resulting in a balanced dataset of $4320$ samples total. Unlike prior work that evaluates CDP authentication in a binary setting by pairing one authentic printer with its counterfeit variants, our formulation includes all printers and counterfeit types jointly. This supports learning of cross-printer signature variations in a unified classification setting.

\inlineheading{Data preparation and augmentation}
To ensure robust evaluation, the dataset is split into $70\%$ training ($504$ samples/class), $10\%$ validation ($72$ samples/class), and $20\%$ testing ($144$ samples/class). To prevent information leakage, the splits are based on template ID, ensuring each template and printed variants appear in only one of the splits. This enforces that the model generalizes to unseen templates during evaluation, which is more realistic and difficult. At both training and inference time, the corresponding binary template \(b\) used to generate each printed CDP is assumed to be available, as it can be retrieved from a secured database in typical deployment settings.

During training, each template--CDP pair is augmented $20$ times via random $512{\times}512$ crops (always), horizontal/vertical flips ($p{=}0.8$), and for printed CDPs only, photometric transforms: Gaussian blur (kernel 3--7), Gaussian noise (var. 0.001--0.005), and brightness/contrast shifts ($\pm0.2$), each with applied probability $p{=}0.7$. Two photometric transforms are applied per image ($p{=}0.7$). See our code repository for full augmentation parameters and implementation details.

\inlineheading{Evaluation criteria} 
We evaluate the model using a multiclass classification setup, where each class corresponds to one of two authentic printers or one of four counterfeit printing scenarios. The authentic classes (\textbf{HP55}, \textbf{HP76}) represent prints from HP Indigo 5500 and HP Indigo 7600, respectively. The counterfeit classes are derived by reprinting estimated templates using either printer.

Evaluation is conducted along two complementary axes:
\begin{enumerate}
    \item \textbf{Classification accuracy}\\
    This measures the model’s ability to correctly identify the source printer class for each printed CDP across all six classes. Results are reported using a normalised confusion matrix, with each row representing the ground truth and each column the predicted class.

    \item \textbf{Authentication performance}\\
    Authentication is formulated as a binary decision: whether the predicted class matches the expected authentic printer. We use the balanced error rate \(P_{\text{err}}\), defined as the average of the false rejection rate (\(P_{\text{miss}}\)) and the false acceptance rate (\(P_{\text{fa}}\)):
    
    \begin{equation}
    \text{P}_{\text{err}} = \frac{ \text{P}_{\text{miss}} + \text{P}_{\text{fa}} }{2}
    \label{eq:auth-eval}
    \end{equation}

\end{enumerate}
This metric equally penalizes both error types and is robust to class imbalance. Here, \(P_{\text{miss}}\) is the proportion of authentic samples incorrectly rejected, and \(P_{\text{fa}}\) is the proportion of counterfeit samples incorrectly accepted.

\begin{table*}[ht]
\centering
\caption{Authentication error rate for method comparison with baselines. Lower is better.}
\label{tab:acc-method-comparison}
\tiny
\resizebox{\textwidth}{!}{%
\begin{tabular}{llccccccccc}
\hline
                               &                             & \multicolumn{1}{l}{$\textbf{P}_{\text{err}}$} & \multicolumn{3}{c}{$\textbf{P}_{\text{miss}}$}                                                 & \multicolumn{5}{c}{$\textbf{P}_{\text{fa}}$}                                                                                                  \\ \hline
                               & \multicolumn{1}{l|}{}       & \multicolumn{1}{l|}{}             & \multicolumn{1}{l}{\textit{HP55}} & \textit{HP76} & \multicolumn{1}{c|}{\textit{Mean}} & \multicolumn{1}{l}{\textit{\textit{HP55\_55}}} & \multicolumn{1}{l}{\textit{HP55\_76}} & \textit{HP76\_76} & \textit{HP76\_55} & \textit{Mean} \\
\multirow{2}{*}{Traditional}   & \multicolumn{1}{l|}{NCC}    & \multicolumn{1}{c|}{0.286}           & 0.292                            & 0.310            & \multicolumn{1}{c|}{ 0.301}            & 0.300                            & 0.361                                    & 0.264                & 0.201                &  0.273               \\
                               & \multicolumn{1}{l|}{SSIM}   & \multicolumn{1}{c|}{0.275}           & 0.264                             & 0.300            & \multicolumn{1}{c|}{0.281}            & 0.292                            & 0.285                                    & 0.292                & 0.210                &  0.269                \\ \hline
\multirow{2}{*}{Deep Learning} & \multicolumn{1}{l|}{{\cite{DBLP:conf/wifs/ChabanPV24}}} & \multicolumn{1}{c|}{0.091}           & 0.125                           & 0.111            & \multicolumn{1}{c|}{0.118}            & 0.097                            & 0.16                                    & \textbf{0.000}                & 0.000                &  0.064               \\
                               & \multicolumn{1}{l|}{Ours}   & \multicolumn{1}{c|}{\textbf{0.023}}           & \textbf{0.049}                            & \textbf{0.035 }            & \multicolumn{1}{c|}{\textbf{0.005}}            & \textbf{0.000}                            & \textbf{0.014}                                    & 0.007                & \textbf{0.000}                & \textbf{0.042}                \\ \hline
\end{tabular}%
}
\end{table*}

\inlineheading{Training pipeline} 
The authentication framework is built using pretrained Stable Diffusion \cite{DBLP:conf/cvpr/RombachBLEO22}. To better preserve structural information in binary templates, the variational autoencoder (VAE) was fine-tuned on template data, as the original model showed limited reconstruction fidelity for high-frequency details. The VAE fine-tuning was performed using the same template images from the dataset splits and with the same training hyperparameters later used for ControlNet training. We then trained the ControlNet components used for image-based conditioning, specifically the parallel branch of the denoising U-Net and its associated image encoder, using a diffusion loss to align the denoising trajectory with the conditioning inputs. All other components, including the main U-Net backbone and CLIP-based text encoder, were kept frozen. Training was conducted in mixed precision (FP16/FP32) on an NVIDIA H100 80GB GPU, with a batch size of $16$ and gradient accumulation of $4$. The optimiser used was AdamW, with a cosine learning rate schedule and linear warmup over the first $500$ steps, starting from an initial learning rate of \(8 \times 10^{-5}\). Training was performed for $100$ epochs. During inference, authentication was performed using $50$ denoising trials per sample to ensure prediction stability.

\subsection{Results and Discussion}
\label{subsec:results-and-discussion}
\inlineheading{Classification accuracy} \Cref{tab:conf-mat-classification} presents the classification accuracy for each class using a confusion matrix. The high diagonal values across the matrix demonstrate strong classification performance. Both authentic classes are consistently and accurately identified, indicating high reliability in recognising genuine CDPs. This is crucial for minimising false rejections in the authentication pipeline.

Counterfeit classes show asymmetric confusion, especially within the same source printer (e.g., HP55 vs HP76). For instance, \textbf{HP76\_55} is misclassified as \textbf{HP76\_76} in $19\%$ of cases. This suggests the model struggles to fully disentangle the latent signatures inherited from the original authentic printer. However, no such confusion occurs across counterfeit classes originating from different source printers, indicating that the model effectively distinguishes CDPs from distinct authentic origins. Since confusion is within the counterfeit domain, we next evaluate its impact on authentication.

\begin{table}[t]
\centering
\caption{Confusion matrix of classification results. Values are in percentages.}
\label{tab:conf-mat-classification}
\footnotesize
\setlength{\tabcolsep}{3pt}
\begin{tabular}{llrrrrrr}
\toprule
\multicolumn{2}{c}{} & \multicolumn{6}{c}{Predicted class (\%)} \\
\cmidrule(lr){3-8}
\multicolumn{2}{c}{} & HP55 & HP76 & HP55\_55 & HP55\_76 & HP76\_76 & HP76\_55 \\
\midrule
\multirow{6}{*}{\rotatebox{90}{Actual class (\%)}} 
& HP55     & \textbf{95} &  5 &  0 &  0 &  0 &  0 \\
& HP76     &  3 & \textbf{97} &  0 &  0 &  0 &  0 \\
& HP55\_55 &  0 &  0 & \textbf{83} & 15 &  1 &  1 \\
& HP55\_76 &  1 &  0 & 14 & \textbf{83} &  1 &  1 \\
& HP76\_76 &  0 &  1 &  1 &  1 & \textbf{89} &  8 \\
& HP76\_55 &  0 &  0 &  2 &  2 & 19 & \textbf{77} \\
\bottomrule
\end{tabular}
\end{table}

\inlineheading{Authentication performance} We evaluate the authentication performance of our proposed method using the balanced error rate obtained from false rejections and false acceptances. Our results are compared against traditional similarity-based baselines and a deep learning approach adapted from~\cite{DBLP:conf/wifs/ChabanPV24}. The results include overall performance ($P_{\text{err}}$), error rates for authentic and counterfeit CDPs ($P_{\text{miss}}$ and $P_{\text{fa}}$), and a per-class analysis of false authentications.

The results are presented in \Cref{tab:acc-method-comparison}. Traditional baselines rely on computing similarity metrics between the candidate CDP and its binary template. We use Normalised Cross-Correlation (NCC) and Structural Similarity Index (SSIM) as representative metrics \cite{DBLP:conf/wifs/BelousovPCTTHV22, Chaban2021MachineLA, DBLP:conf/wifs/YadavTTF19, 10715309}. For each class, we determine a similarity threshold from the validation set. During testing, samples are authenticated based on whether their similarity score exceeds the threshold for their class. While conceptually simple, these methods are highly sensitive to printer-specific variations and fail to model inter-class differences effectively. Their performance is modest across all classes, with similar error values; neither method shows a clear advantage in correctly accepting authentic samples or rejecting counterfeits, with errors generally ranging between $0.269$ and $0.300$.

Chaban \etal\cite{DBLP:conf/wifs/ChabanPV24} trained a Pix2Pix model\cite{DBLP:conf/cvpr/IsolaZZE17} to synthesize a CDP from the input binary template, then compared this synthesis to the candidate CDP using the Pearson Correlation Coefficient. In the original setup, separate models were trained per printer, and counterfeit data was not considered. For a fair comparison, we retrained their model using their open-source code, incorporating both authentic and counterfeit classes in a unified training pipeline. This improved its generalizability, achieving as little as $P_{fa} = 0.064$ for falsely accepting counterfeit classes. However, its performance remains inconsistent on authentic samples, with a miss rate of $0.125$ for the \textbf{HP55} class.

Our method, conditioned jointly on printer identity and CDP images, outperforms all baselines, achieving a low overall error rate $P_{err}$ of $\textbf{0.023}$. All counterfeit classes are reliably rejected, including those most prone to confusion in the classification step such as \textit{HP55\_55} and \textit{HP76\_55}. On the authentic side, our method achieves the lowest miss rates across both classes, with $P_{miss} = 0.005$ for all authentic CDPs. Notably, the classes that were most challenging for the baseline methods, such as \textbf{HP55} and \textbf{HP55\_76}, show improved error rates under our model. This suggests that conditioning on printer identity during training enables the model to better capture intra-class variability and learn more discriminative representations, particularly in challenging cases.

Although some counterfeit classes exhibit internal confusion during classification, the authentication decision remains robust, with low $P_{fa}$ across all counterfeit types.

This robustness comes from the authentication strategy, which verifies whether the predicted printer identity for a candidate CDP matches the known authentic printer. Consequently, even if a counterfeit is misclassified as another counterfeit class, it is still rejected unless it is erroneously attributed to the correct authentic printer, a scenario that remains rare due to the difficulty of replicating fine-grained, printer-specific structural and spatial features.

\inlineheading{Generalisation to unseen counterfeits}
To assess the model’s robustness against counterfeit types not observed during training, we simulate a realistic deployment scenario where only a subset of forgeries is known. Specifically, we train on authentic CDPs printed with the HP Indigo 5500 and HP Indigo 7600 (HP55, HP76), along with two known counterfeit classes produced by estimating templates from these printers and reprinting them on the same device (HP55\_55, HP76\_76). At test time, we evaluate on:
\begin{enumerate}[label=(\roman*), itemsep=1pt, topsep=2pt]
\item \textbf{Authentic classes:} HP55, HP76
\item \textbf{Known counterfeits:} HP55\_55, HP76\_76
\item \textbf{Unseen counterfeits:} HP55\_76, HP76\_55
\end{enumerate}

This setup tests the model’s ability to reject counterfeits created using printing paths not encountered during training. \Cref{tab:unseen_counterfeits} shows that the model achieves perfect rejection rates ($P_{fa} = 0.000$) on both known and unseen counterfeit types, with a balanced error rate of $P_{err} = 0.012$. These results confirm that the conditioning mechanism enables generalisation beyond the counterfeit strategies seen during training by capturing underlying printer-specific characteristics.

\begin{table}[htbp]
\centering
\caption{Authentication performance on known and unseen counterfeit types. Lower is better.}
\tiny
\label{tab:unseen_counterfeits}
\resizebox{\columnwidth}{!}{%
\begin{tabular}{llll}
\hline \textbf{Counterfeit Type}
                                        & $\textbf{P}_{\text{err}}$                    & $\textbf{P}_{\text{miss}}$                   & $\textbf{P}_{\text{fa}}$ \\ \hline
\multicolumn{1}{l|}{Known counterfeit}  & \multicolumn{1}{l|}{0.012} & \multicolumn{1}{l|}{0.024} & 0.000  \\ 
\multicolumn{1}{l|}{Unknown counterfeit} & \multicolumn{1}{l|}{0.012} & \multicolumn{1}{l|}{0.024} & 0.000  \\ \hline
\end{tabular}%
}
\end{table}

\inlineheading{Binary Template Reconstruction} 
To validate the necessity of VAE fine-tuning, we compared reconstruction quality of binary templates using mean squared error (MSE) between the input and reconstructed outputs under two settings. As shown in \Cref{tab:recon_mse}, the pretrained VAE yielded a high average error $(0.581)$, indicating limited structural preservation. Fine-tuning the VAE on binary templates significantly improved reconstruction fidelity, reducing the error to $(0.075)$ and confirming its suitability for encoding high-frequency, pixel-critical content.

\begin{table}[h]
\centering
\caption{Mean squared reconstruction error between original and reconstructed binary templates under different configurations.}
\small
\label{tab:recon_mse}
\begin{tabular}{l c}
\toprule
\textbf{Configuration} & \textbf{Avg. MSE (↓)} \\
\midrule
Pretrained VAE & 0.581 \\
\emph{Fine-tuned VAE} & \textbf{0.075} \\
\bottomrule
\end{tabular}
\end{table}

\inlineheading{Ablation study}
To assess the contribution of different inputs to the model’s authentication performance, we conduct an ablation study varying two key components: (i) the presence of the binary template, and (ii) how printer identity is represented.  Results are summarized in \Cref{tab:template_ablation} and \Cref{tab:printer_identity_ablation}.

\Cref{tab:template_ablation} shows the impact of removing the binary template. When the template is omitted and replaced by the printed CDP image as input to the architecture, the model relies solely on the printed output and the class label. This leads to a substantial drop in performance, indicating that structural alignment between the template and printed output is essential for capturing fine-grained printer signatures. Without the template as a reference, the model struggles to extract discriminative spatial features, resulting in high error rates.

\Cref{tab:printer_identity_ablation} evaluates different strategies for representing printer identity. When printers are represented by index numbers rather than textual descriptions, performance similarly degrades. Notably, this setup results in a high miss rate for authentic samples but relatively low false acceptances, suggesting a strong bias toward predicting samples as counterfeit. This indicates that numerical indices fail to capture the semantic distinctions between printer classes, limiting the model’s ability to recognise valid CDPs.

As a second approach to printer identity representation, we tested a configuration where each CDP is conditioned on a natural language description of the typical visual appearance of printed samples from that printer. For example, HP55 samples are described as \textit{``A high-contrast black and white Data Matrix symbol with dense black module clusters, darkened center region, and a coarse, grainy texture throughout the square grid''}. This method outperforms index-based conditioning, particularly in rejecting counterfeits. However, authentic acceptance is less consistent, suggesting that appearance-based descriptions lack the specificity needed to reliably guide authentication.

Together, these results highlight the importance of jointly conditioning on both the binary template and semantically meaningful printer identity. Each plays a distinct role: the template provides structural cues aligned with the original CDP layout, while printer identity embeds class-level semantics necessary to distinguish subtle differences in output characteristics across printers. 

\inlineheading{Reproducibility}
To reproduce our experiments, you can refer to our public repository for scripts and models.

\begin{table}[htbp]
\centering
\caption{Authentication error with and without template input. Lower values indicate better performance.}
\tiny
\label{tab:template_ablation}
\resizebox{\columnwidth}{!}{%
\begin{tabular}{llll}
\hline
\textbf{Input Configuration} & $\textbf{P}_{\text{err}}$ & $\textbf{P}_{\text{miss}}$ & $\textbf{P}_{\text{fa}}$ \\
\hline
\multicolumn{1}{l|}{No Template Input}                       & \multicolumn{1}{l|}{0.660}           & \multicolumn{1}{l|}{0.430}            & 0.200 \\
\multicolumn{1}{l|}{\emph{With Template Input}}                       & \multicolumn{1}{l|}{\textbf{0.023}}  & \multicolumn{1}{l|}{\textbf{0.042}}   & \textbf{0.005} \\ \hline
\end{tabular}%
}
\end{table}
\begin{table}[htbp]
\centering
\caption{Authentication error comparing different printer identity representations. Lower values indicate better performance.}
\tiny
\label{tab:printer_identity_ablation}
\resizebox{\columnwidth}{!}{%
\begin{tabular}{llll}
\hline
\textbf{Representation} & $\textbf{P}_{\text{err}}$ & $\textbf{P}_{\text{miss}}$ & $\textbf{P}_{\text{fa}}$ \\
\hline
\multicolumn{1}{l|}{Printer Index based}                     & \multicolumn{1}{l|}{0.521}           & \multicolumn{1}{l|}{0.962}            & 0.082 \\
\multicolumn{1}{l|}{Appearance-based}           & \multicolumn{1}{l|}{0.035}           & \multicolumn{1}{l|}{0.063}            & \textbf{0.001} \\
\multicolumn{1}{l|}{\emph{Printer Information}} & \multicolumn{1}{l|}{\textbf{0.023}}  & \multicolumn{1}{l|}{\textbf{0.042}}   & 0.005 \\ \hline
\end{tabular}%
}
\end{table}

\section{Conclusion}
\label{sec:conclusion}
We presented a novel authentication framework for Copy Detection Patterns (CDPs) that integrates the binary template, printed CDP, and printer identity into a unified classification pipeline based on diffusion modelling. Unlike prior approaches that rely on template similarity or generative synthesis, we reformulate authentication as a per-printer classification task. This allows the model to learn discriminative printer signatures and supports generalisation across different printers.

A key technical contribution of this work is the extension of the ControlNet architecture beyond its generative role to support classification. By modifying the reverse diffusion process to perform class-specific noise prediction, we enable printer identification via minimum reconstruction error. This leverages semantic (text) and spatial (image) conditioning to capture global printer characteristics and localised print artifacts.
 
Experiments on the Indigo 1×1 Base dataset demonstrate the effectiveness of our approach in advancing CDP authentication, particularly through the use of semantically rich representations of printer identity that enhance model performance and generalisation. While these results show strong promise, the current experiments are limited to a single scanner configuration. Future work will explore more diverse acquisition settings, increased variability in printers, and investigate training strategies that do not require counterfeit examples.

\section{Acknowledgements}
\label{sec:acknowledgements}
This work was financed by the French National Research Agency (ANR), project TRUSTIT referenced under ANR-23-CE39-0002-01

{\small
\bibliographystyle{ieee_fullname}
\bibliography{egbib}
}

\end{document}